\documentclass[10pt,twocolumn,letterpaper]{article}

\usepackage[pagenumbers]{cvpr} 

\definecolor{cvprblue}{rgb}{0.21,0.49,0.74}
\usepackage[pagebackref,breaklinks,colorlinks,allcolors=cvprblue]{hyperref}
\usepackage{amsmath}
\usepackage{amssymb}
\usepackage{booktabs}
\usepackage{graphicx}
\usepackage{algorithm}
\usepackage[noend]{algpseudocode}
\algrenewcommand\algorithmicrequire{\textbf{Input:}}
\algrenewcommand\algorithmicensure{\textbf{Output:}}

\def\confName{CVPR}
\def\confYear{2027}

\title{PRC4D: Physical Relation-Constrained Persistent 4D Reconstruction}

\author{
Junho Kim$^{*}$ \qquad
Hyunyoung Choi$^{*}$ \qquad
Jiseok Kim \qquad
Seongwon Lee$^{\dagger}$ \\
\\
Kookmin University \\
{\tt\small
\{jhk00,kellychy,inplus1130,sungonce\}@kookmin.ac.kr
}
}

\begin{document}
\twocolumn[{
    \renewcommand\twocolumn[1][]{#1}%
    \maketitle
     \includegraphics[width=\textwidth]{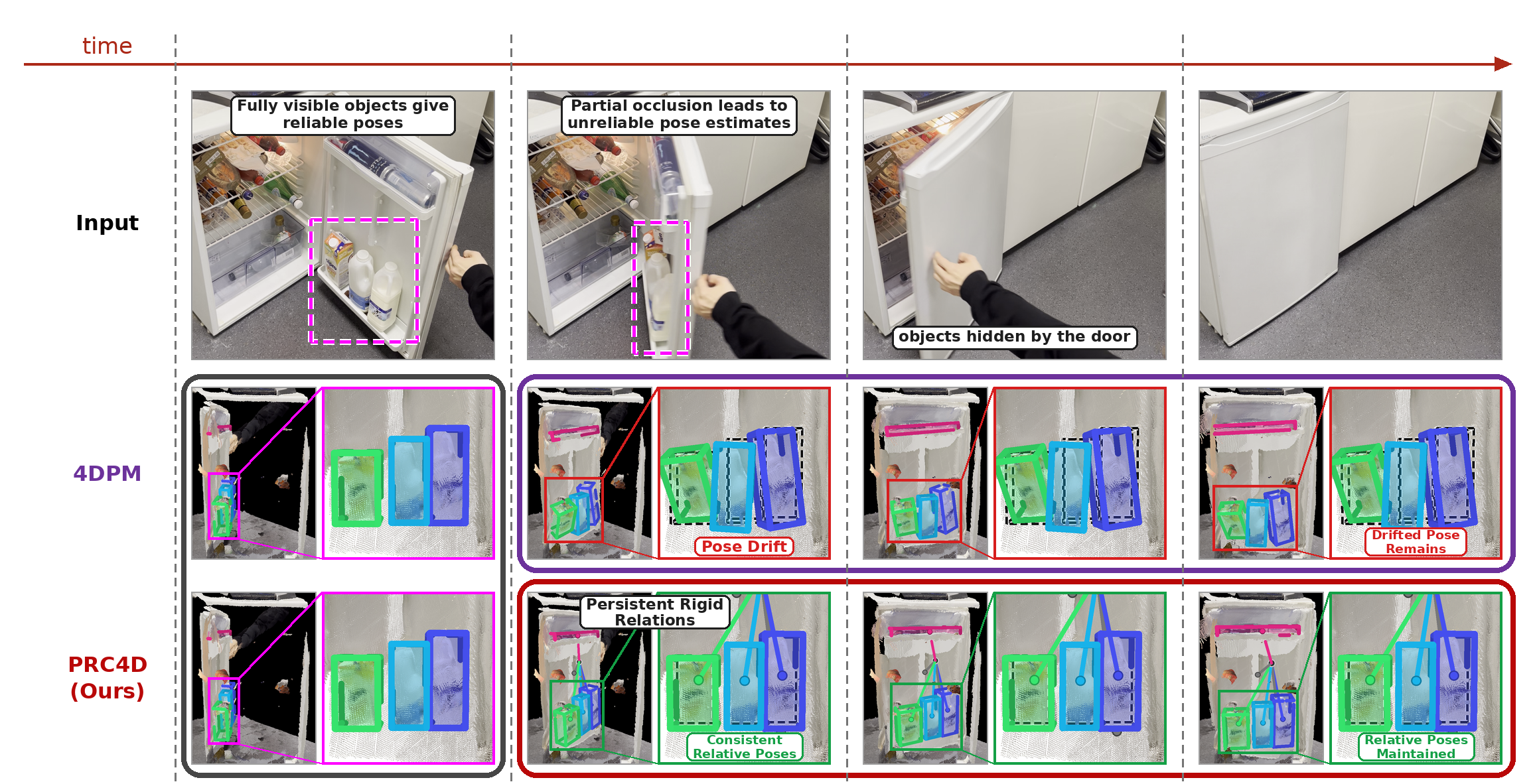}
      \captionof{figure}{
        Our method (\textbf{PRC4D}) infers persistent rigid relations
        to maintain spatial consistency through occlusion.
        As the refrigerator door closes, objects become
        progressively occluded (\textbf{top row}).
        While 4DPM preserves their geometry, its estimated
        poses drift (\textbf{middle row}).
        In the final frame, PRC4D maintains the correct
        relative configurations of fully occluded objects
        by propagating motion through verified rigid relations
        (\textbf{bottom row}).
      }
  \label{fig:teaser}
  \vspace{2em}
}]

\begingroup
\renewcommand{\thefootnote}{}
\footnotetext{
$^{*}$ Co-first authors.
\quad $^{\dagger}$ Corresponding author.
}
\endgroup

\begin{abstract}


A faithful 4D reconstruction should capture both the shape and motion of objects, even when they disappear from view. Existing methods preserve previously observed geometry and extend motion through occlusion using local spatial and motion associations. However, treating short-term contact or shared motion as sufficient evidence of persistent rigidity can introduce incorrect motion constraints and distort overall reconstruction. To address this limitation, we present PRC4D, a physical relation-constrained framework that identifies reliable rigid relations and uses them to maintain spatial consistency through occlusion. PRC4D combines 3D contact and relative motion consistency with sequence-level verification to distinguish persistent couplings from transient interactions. The resulting graph guides observed-pose refinement
and motion propagation from visible to occluded
components without altering reconstructed geometry. Experiments on diverse real-world multi-object sequences and controlled occlusion simulations demonstrate improved reconstruction accuracy, reduced relative pose errors, and more consistent object configurations under changing visibility.
\end{abstract}    
\section{Introduction}

Monocular videos provide only partial observations of dynamic scenes, leaving the motion of occluded objects uncertain.
Recent advances in monocular 4D reconstruction have improved geometric recovery and motion estimation through visual correspondences~\cite{doersch2023tapir,karaev2024cotracker,wang2024searaft} and structured motion representations~\cite{zhang2022structuremotion,zhang2025monst3r,li2025megasam}.
In particular, 4D Primitive-Mâché (4DPM)~\cite{mazur2026primitive} accumulates previously observed geometry as 3D primitives and extends their motion through occlusion, enabling persistent reconstruction beyond the currently visible scene.

However, approaches that infer occluded motion from local spatial and motion associations can mistake transient interactions for persistent rigid couplings.
Components may remain in contact while undergoing relative motion, as in articulated structures~\cite{tresadern2005articulated,liu2023paris,jiang2022ditto}, or briefly move together before separating.
Treating these associations as fixed rigid relations can introduce incorrect motion constraints and distort relative object poses.
The central challenge is therefore not simply to propagate object motion, but to determine which physical relations provide reliable constraints over time.

To address this limitation, we present PRC4D, a physical relation-constrained framework for persistent 4D reconstruction.
We infer candidate rigid couplings from 3D contact and relative motion consistency, then verify their persistence using sequence-level evidence and graph-level consistency.
The verified relations form a temporal physical-relation graph that identifies which reconstructed components remain rigidly coupled.

The verified graph constrains both observed and occluded motion.
For observed components, we refine their poses to reduce deviations from the relative configurations of their rigid groups.
When a component becomes occluded, we propagate motion from a visible member of the same group, maintaining the relative configuration as visibility changes (Fig.~\ref{fig:teaser}).

We evaluate PRC4D on diverse real-world multi-object sequences and demonstrate improved reconstruction accuracy.
To further assess its robustness to occlusion, we construct OmniGibson-Occlusion, a controlled synthetic benchmark that provides ground-truth geometry and motion throughout both visible and occluded intervals.
PRC4D achieves the highest mean reconstruction F-score among the compared methods on both benchmarks, with controlled occlusion experiments additionally showing reduced relative pose errors and more consistent object configurations.

Our contributions are summarized as follows:
\begin{itemize}
    \item We introduce a temporal physical-relation graph that distinguishes persistent rigid couplings from transient interactions by combining geometric and motion evidence with sequence-level verification.

    \item We develop a relation-constrained reconstruction framework that uses the same verified relations to refine observed component poses and propagate motion through occlusion.

    \item We construct OmniGibson-Occlusion, a synthetic benchmark with ground-truth geometry and motion throughout occlusion, enabling controlled evaluation of reconstruction robustness under changing visibility.

    \item We demonstrate improved reconstruction accuracy on real-world and synthetic benchmarks, with controlled occlusion experiments showing reduced relative pose errors and more consistent object configurations.
\end{itemize} 
\section{Related Work}

\paragraph{Monocular 4D Reconstruction.}
Recent advances in monocular 4D reconstruction have improved the recovery of dynamic scene geometry and motion through geometric priors~\cite{wang2024dust3r,leroy2024mast3r,wang2025vggt,wang2026pi3,murai2025mast3rslam}, temporal correspondences~\cite{doersch2023tapir,karaev2024cotracker,harley2025alltracker, wang2023omnimotion}, and structured motion representations~\cite{zhang2025monst3r,zhou2026page4d,sucar2025dynamicpointmaps, sucar2026vdpm, chen2025easi3r, matsuki2025_4dtam}. Optimization-based approaches, such as Shape of Motion~\cite{wang2025shapeofmotion} and C4D~\cite{wang2025c4d}, integrate long-range correspondences and geometric constraints to recover coherent 3D motion. Meanwhile, feed-forward methods, including POMATO~\cite{zhang2025pomato}, St4RTrack~\cite{feng2025st4rtrack}, and D4RT~\cite{zhang2026d4rt}, jointly estimate spatiotemporal geometry and motion from video. However, temporal correspondences and motion regularization do not necessarily enforce constant relative transformations between distinct components that move rigidly together. Our method addresses this by inferring persistent rigid relations from reconstructed geometry and motion, and using them as explicit cross-component SE(3) constraints for pose refinement.

\begin{figure*}[h!]
    \centering
    \includegraphics[width=\linewidth]{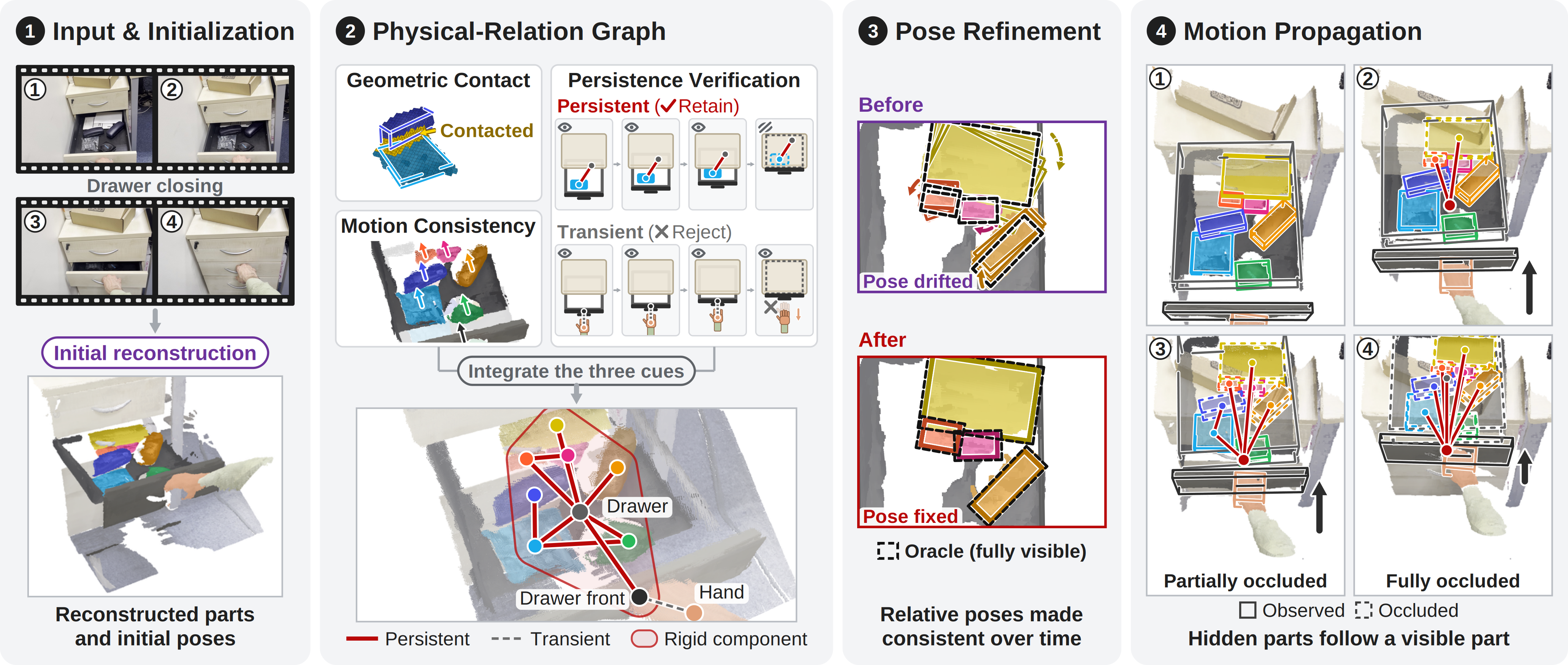}
    \caption{
    \textbf{Overview of PRC4D.}
    Geometric contact and relative motion provide
    candidate rigid relations, which are verified
    through sequence-level persistence reasoning.
    The resulting graph refines observed component
    poses and guides motion propagation through occlusion.
    }
    \vspace{-12pt}
    \label{fig:method}
\end{figure*}

\paragraph{Occlusion-Aware and Persistent 4D Reconstruction.}
Occlusion-aware 4D reconstruction has been explored through spatiotemporal reasoning, geometric completion, and persistent scene representations~\cite{luiten2024dynamic3dgaussians, wang2025cut3r, huang2025vivid4d}. Recent approaches, such as Complet4R~\cite{wang2026complet4r} and COM4D~\cite{gokmen2026com4d}, leverage information across observations to reconstruct geometry beyond the current view and recover compositional 4D scenes. 4DPM~\cite{mazur2026primitive} further enables persistent reconstruction by accumulating previously observed geometry and extending object motion through occlusion using spatial contact and motion similarity. However, such local associations do not necessarily establish persistent rigid couplings across a sequence. Our method addresses this limitation through sequence-level relation verification and relation-constrained pose refinement and motion propagation.

\paragraph{Rigid Motion Constraints and Relational Reconstruction.}
Geometric and relational constraints provide additional structure for reconstructing dynamic scenes from ambiguous visual observations~\cite{costeira1995multibody,kumar2021superpixelsoup,ost2021neural_scene_graphs}. MoSca~\cite{lei2025mosca} represents scene deformation using sparse motion scaffolds, enforcing local rigidity and smoothness to recover coherent motion. Beyond deformation regularization, CARI4D~\cite{xie2026cari4d} reasons about human-object contacts and incorporates them into joint pose optimization to improve spatial and temporal alignment. Relational reasoning has also been explored in HAT-4D~\cite{li2026hat4d}, which constructs an interaction knowledge graph to guide multi-object scene composition and memory-guided 4D generation. These approaches demonstrate the value of motion and interaction constraints for dynamic reconstruction. Our work focuses on verifying the temporal persistence of rigid relations between reconstructed scene components and using them to constrain both observed and occluded motion.

\section{Method}

\subsection{Problem Formulation and Overview}

Given a monocular RGB video
$\mathcal{I}=\{I_t\}_{t=1}^{T}$,
our goal is to reconstruct a dynamic scene
while preserving coherent spatial configurations
across changing visibility, including occlusion.

We obtain an initial primitive-based reconstruction
using 4D Primitive-M\^ach\'e (4DPM)~\cite{mazur2026primitive}, which
recovers scene geometry and estimates component
motion from visual correspondences.
Let $\mathcal{O}=\{O_i\}_{i=1}^{N}$ denote the
reconstructed scene components, each associated
with 3D geometry and estimated poses
$\tilde{\mathbf{T}}_i^t\in SE(3)$ at observed
keyframes in a common coordinate frame.
Although the reconstructed geometry persists,
independently estimated component poses may
violate the relative configurations of components
that move rigidly together.

To address this inconsistency, we construct a
temporal physical-relation graph
$\mathcal{G}=(\mathcal{V},\mathcal{E})$,
where nodes represent reconstructed components
and edges encode inferred persistent rigid relations.
For a rigidly coupled parent-child pair
$(p,c)\in\mathcal{E}$, we seek poses satisfying
\begin{equation}
(\mathbf{T}_p^t)^{-1}\mathbf{T}_c^t
\approx \bar{\mathbf{T}}_{pc},
\label{eq:relative_constraint}
\end{equation}
where $\bar{\mathbf{T}}_{pc}\in SE(3)$ denotes
their reference relative transformation.

Our framework infers persistent relations from
geometric and temporal evidence, then uses the
resulting graph to refine observed component
poses and propagate motion to occluded components
through their visible rigid partners.
The refined motions are applied to the original
3D primitives without modifying their geometry (Fig.~\ref{fig:method}).

\subsection{Temporal Physical-Relation Graph}
\label{sec:temporal_graph}

\paragraph{Geometric Contact Evidence.}
We first identify potential rigid couplings using
the accumulated 3D geometry of reconstructed components.
Candidate pairs are selected based on overlap between
expanded oriented bounding boxes (OBBs), followed by
geometric contact evaluation using voxelized surfaces
at jointly observed keyframes.
Contact is measured by the proportion of surface voxels
within a spatial neighborhood of the other component.
Temporal hysteresis aggregates noisy observations
into continuous contact intervals.
These intervals provide evidence of physical contact,
but do not by themselves establish rigid coupling.

\paragraph{Rigid Motion Consistency.}
We further examine whether contacting components
exhibit consistent motion.
Following the 4DPM convention, we define the motion
increment of component $i$ as
$\mathbf{D}_i^t =
(\mathbf{U}_i^t)^{-1}\mathbf{U}_i^{t-1}$,
where $\mathbf{U}_i^t$ denotes its original transform.
For a candidate relation from child $i$ to parent $j$,
let $(\boldsymbol{\tau}_{ij}^t,
\boldsymbol{\phi}_{ij}^t)$ denote the translational
and rotational components of
$\operatorname{Log}_{SE(3)}
((\mathbf{D}_j^t)^{-1}\mathbf{D}_i^t)$.
We compute the normalized motion discrepancy as
\begin{equation}
\epsilon_{ij}^t =
\max\left(
\frac{\|\boldsymbol{\tau}_{ij}^t\|}
{\max(\alpha s_i,b)},
\frac{\|\boldsymbol{\phi}_{ij}^t\|}{2\beta}
\right),
\label{eq:motion_consistency}
\end{equation}
where $s_i$ is the component size, $\alpha$
and $b$ control translation tolerance, and
$\beta$ controls angular tolerance.

Motion consistency is evaluated only at transitions
with measurable motion, since shared stillness
does not provide evidence of rigid coupling.
Contact intervals with consistent informative
motion and bounded relative translation drift
form candidate rigid relations.

\paragraph{Sequence-Level Persistence Reasoning.}
Local contact and motion consistency provide
evidence of rigid coupling, but may also arise
from transient interactions.
We therefore verify candidate relations across
the complete sequence by examining evidence
that contradicts persistent rigidity.

For each candidate pair $(i,j)$, we examine
contradictory evidence from separated fixed-contact
episodes and independent motion beyond the
supporting interval.
We additionally measure relative translation drift
at jointly observed keyframes as
\begin{equation}
d_{ij}^{t} =
\frac{\|\mathbf{p}_{ij}^{t}-
\mathbf{p}_{ij}^{\mathrm{ref}}\|_2}{s_i},
\label{eq:relation_drift}
\end{equation}
where
$\mathbf{p}_{ij}^{t} =
\operatorname{trans}((\tilde{\mathbf{T}}_j^t)^{-1}
\tilde{\mathbf{T}}_i^t)$
denotes the relative translation, and
$\mathbf{p}_{ij}^{\mathrm{ref}}$ is its median
over the early half of the final supporting interval.
When this interval extends to the last joint
observation, sustained terminal drift indicates
a possible violation of fixed rigidity.
Missing observations alone do not invalidate
a relation, since occlusion does not imply
relation termination.

We further exploit graph-level consistency to
resolve ambiguities remaining after pairwise
temporal verification.
Relations with sufficient motion evidence and
no temporal contradictions first establish
trusted rigid groups.
Remaining candidates are evaluated against
these groups, allowing consistent relations
to reinforce or extend existing groups while
rejecting connections between independently
established trusted groups.

Relations with separated fixed-contact episodes
are classified as transient.
For candidates exhibiting independent motion
or terminal drift, we examine redundant support
from distinct neighbors within the same trusted
group.
Sufficient support can retain these uncertain
relations, while unsupported candidates are rejected.
This prevents isolated noisy observations from
breaking otherwise consistent rigid assemblies.

The resulting persistent rigid-relation graph
provides shared constraints for pose refinement
and motion propagation through occlusion.

\begin{table*}[t]
\centering
\caption{
\textbf{Quantitative comparison on our OmniGibson-Occlusion Benchmark.}
We report F-score ($\uparrow$) for each of the six scenes,
along with mean F-score, precision, and recall.
Best and second-best results among non-oracle
methods are highlighted in bold and underlined,
respectively.
The last-view oracle uses only the final-frame
pointmap and is excluded from the ranking.
}
\label{tab:synthetic}

\setlength{\tabcolsep}{4pt}
\renewcommand{\arraystretch}{0.92}

\resizebox{0.93\textwidth}{!}{%
\begin{tabular}{lcccccc|ccc}
\toprule
Method
& Fridge1 & Fridge2
& Drawer1 & Drawer2
& Tray & Storage Box
& Mean & Precision & Recall \\
\midrule

$\pi^3$ (Last-view Oracle)
& 0.1913 & 0.1313
& 0.1306 & 0.1187
& 0.1178 & 0.1469
& 0.1394 & 0.7910 & 0.0769 \\

\midrule

$\pi^3$
& 0.1812 & 0.2185
& 0.2327 & 0.2298
& 0.3769 & 0.2044
& 0.2406 & 0.2111 & 0.2815 \\

St4RTrack
& 0.1291 & 0.0624
& 0.1861 & 0.1574
& 0.1144 & 0.1830
& 0.1387 & 0.1300 & 0.1500 \\

POMATO
& 0.3316 & 0.2515
& 0.2251 & 0.2793
& 0.1042 & 0.2755
& 0.2445 & 0.2352 & 0.2564 \\

TraceAny
& 0.1625 & 0.1852
& 0.1417 & 0.0927
& 0.0618 & 0.0867
& 0.1218 & 0.1159 & 0.1297 \\

D4RT
& 0.3156 & 0.3625
& 0.3025 & 0.2351
& 0.1382 & 0.3313
& 0.2809 & 0.2644 & 0.3015 \\

4DPM
& \underline{0.5835} & \underline{0.4872}
& \underline{0.7687} & \underline{0.6704}
& \underline{0.3772} & \underline{0.6314}
& \underline{0.5864} & \underline{0.5922}
& \underline{0.5843} \\

\midrule

\textbf{PRC4D (Ours)}
& \textbf{0.5982} & \textbf{0.5077}
& \textbf{0.7950} & \textbf{0.6802}
& \textbf{0.3952} & \textbf{0.6790}
& \textbf{0.6092} & \textbf{0.6083}
& \textbf{0.6140} \\

\bottomrule
\end{tabular}%
}

\vspace{-3mm}
\end{table*}

\begin{figure*}[t]
    \centering
    \vspace*{7mm}
    \includegraphics[width=1.0\textwidth]{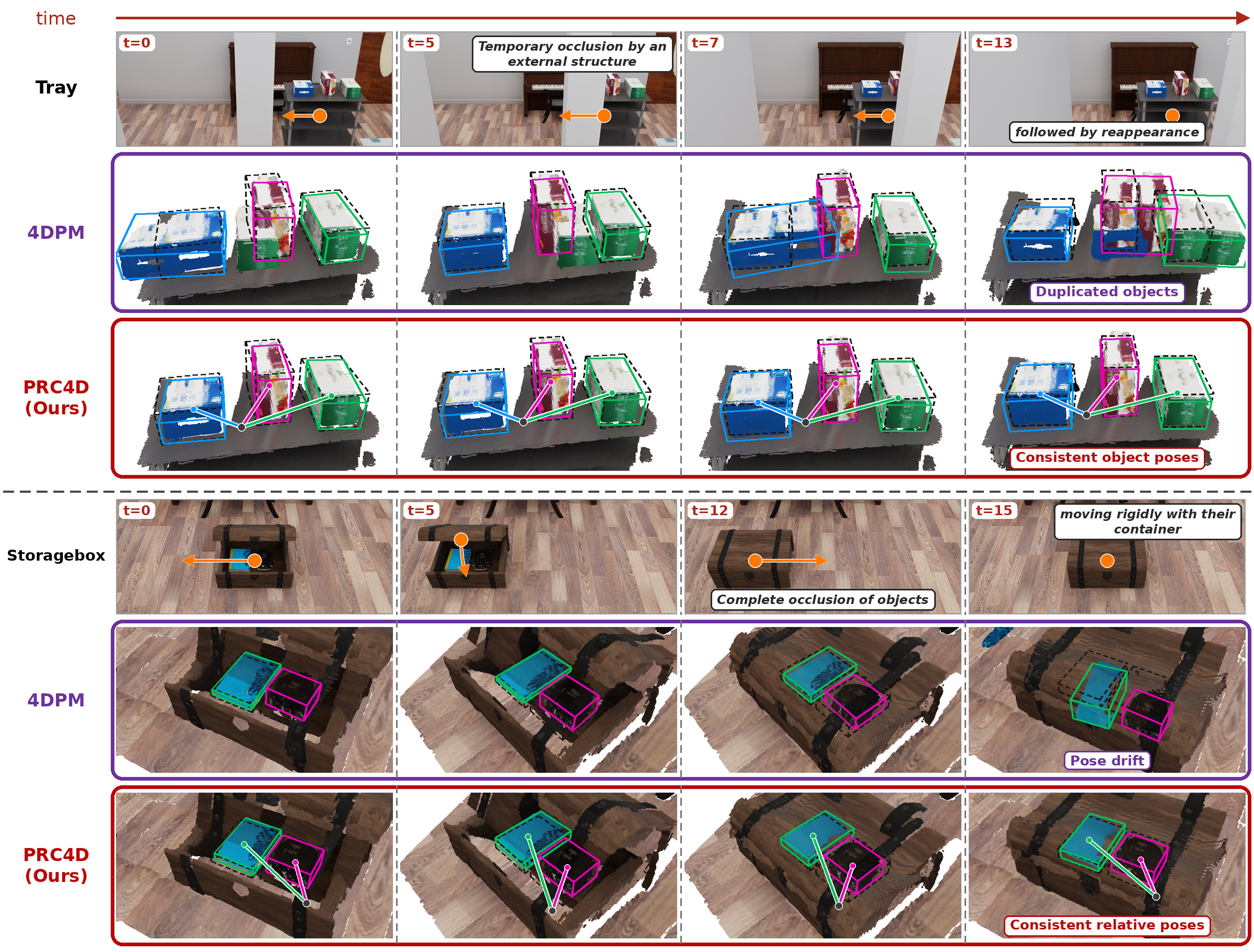}

    \caption{
    \textbf{Persistent 4D reconstruction under occlusion.}
    We compare 4DPM and ours on \textbf{Tray} (top),
    featuring temporary occlusion and reappearance,
    and \textbf{Storage Box} (bottom), where objects
    move with their container under complete occlusion.
    Our persistent physical-relation constraints
    prevent inconsistent object configurations and
    reduce pose drift observed in 4DPM.
    }
    \label{fig:synthetic_qualitative}
\end{figure*}

\subsection{Graph-Constrained Pose Refinement}
\label{sec:pose_refinement}

\begin{table*}[t]
\centering
\caption{
\textbf{Quantitative comparison on the multi-object reconstruction benchmark.}
We report F-score ($\uparrow$) for each scene,
along with mean F-score, precision, and recall.
Best and second-best results among non-oracle
methods are highlighted in bold and underlined,
respectively.
The last-view oracle uses only the final-frame
pointmap and is excluded from the ranking.
}
\label{tab:multiobject}

\setlength{\tabcolsep}{4pt}
\renewcommand{\arraystretch}{0.92}

\resizebox{0.93\textwidth}{!}{%
\begin{tabular}{lccccccc|ccc}
\toprule
Method
& BallPush & BoxSpin
& MultiObj1 & MultiObj2
& PanStir & Spin1 & Spin2
& Mean & Precision & Recall \\
\midrule

$\pi^3$ (Last-view Oracle)
& 0.4011 & 0.5228 & 0.6438
& 0.8273 & 0.1811
& 0.4727 & 0.5121
& 0.5087 & 0.8818 & 0.3713 \\

\midrule

$\pi^3$
& 0.2288 & 0.6005 & \underline{0.6809}
& 0.6881 & 0.1630
& 0.3991 & 0.5933
& 0.4791 & 0.3626 & 0.7391 \\

St4RTrack
& 0.2688 & 0.6343 & 0.3625
& 0.6223 & 0.4394
& 0.5101 & 0.5198
& 0.4796 & 0.3533 & 0.7599 \\

POMATO
& 0.0745 & 0.4603 & 0.6087
& 0.6122 & 0.0650
& 0.2187 & 0.4473
& 0.3553 & 0.2592 & 0.5907 \\

TraceAny
& 0.2927 & 0.5500 & 0.5323
& 0.4253 & 0.3544
& 0.2882 & 0.4938
& 0.4195 & 0.3493 & 0.5338 \\

D4RT
& 0.6485 & 0.7070 & 0.6135
& 0.6525 & \textbf{0.6586}
& 0.5603 & 0.6631
& 0.6433 & 0.5304 & 0.8239 \\

4DPM
& \textbf{0.7986}
& \underline{0.9218}
& \textbf{0.8203}
& \underline{0.8939}
& 0.6215
& \underline{0.8482}
& \underline{0.6732}
& \underline{0.7968}
& \underline{0.7269}
& \underline{0.8931} \\

\midrule

\textbf{PRC4D (Ours)}
& \underline{0.7787}
& \textbf{0.9292}
& \textbf{0.8203}
& \textbf{0.9560}
& \underline{0.6305}
& \textbf{0.8496}
& \textbf{0.7100}
& \textbf{0.8106}
& \textbf{0.7461}
& \textbf{0.8985} \\

\bottomrule
\end{tabular}%
}

\vspace{-3mm}
\end{table*}

\begin{figure*}[t]
    \centering
    \vspace*{2mm}
    \includegraphics[width=0.90\textwidth]{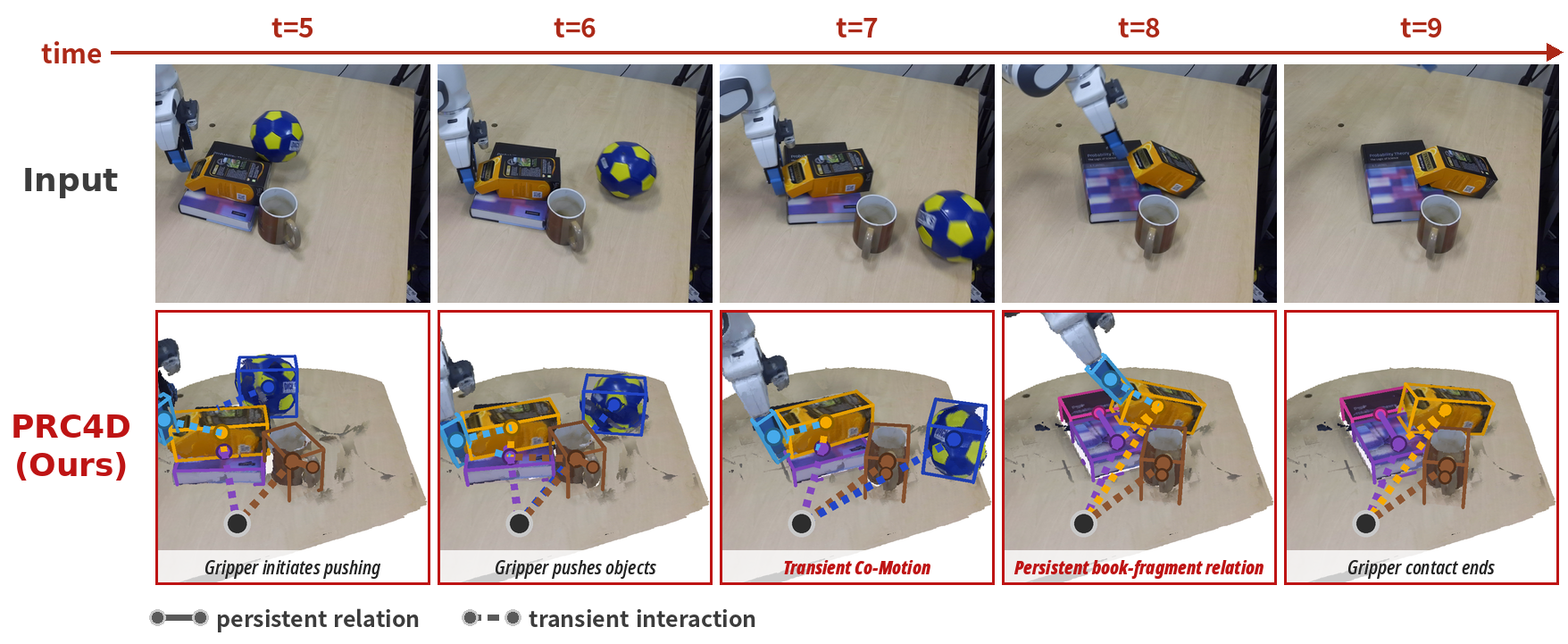}

    \caption{
    \textbf{Persistent and transient physical relations.}
    In \textbf{MultiObj2}, PRC4D distinguishes
    persistent rigid relations between reconstructed
    book fragments from transient contacts and
    co-motion during robot interactions.
    Solid and dashed edges indicate persistent and
    transient relations, respectively.
    }
    \label{fig:multiobject_relations}
\end{figure*}

\paragraph{Reference Relative Configuration.}
For each persistent rigid relation between parent $p$
and child $c$, we estimate a reference transformation
$\bar{\mathbf{T}}_{pc}$ by robustly averaging
$(\tilde{\mathbf{T}}_p^t)^{-1}\tilde{\mathbf{T}}_c^t$
over jointly observed keyframes within its supporting
interval. We employ Huber-weighted averaging in
$SE(3)$ to reduce the influence of noisy estimates,
establishing a reference configuration for
subsequent refinement.

\paragraph{Relation-Constrained Optimization.}
We refine observed component poses by enforcing
consistent relative configurations while preserving
the initial visual pose estimates.
At keyframe $t$, let $\mathcal{E}_t$ denote
the active persistent relations whose endpoints
are jointly observed, and $\mathcal{C}_t$ their
child components.
We optimize the poses of $\mathcal{C}_t$,
keeping other components fixed to their initial
estimates. A parent remains anchored unless
it is itself a child of another active relation.

We define the scale-normalized residual
$\mathbf{r}_i(\mathbf{A},\mathbf{B}) =
\mathbf{S}_i^{-1}\mathbf{q}(\mathbf{A}^{-1}\mathbf{B})$,
where $\mathbf{q}$ stacks the relative translation
and $SO(3)$ rotation logarithm, and $\mathbf{S}_i$
balances their scales. We solve
\begin{equation}
\begin{aligned}
\min_{\{\mathbf{T}_i^t\}_{i\in\mathcal{C}_t}}
&\sum_{i\in\mathcal{C}_t}
\mathcal{L}\left(
\sqrt{\lambda_o}\,
\mathbf{r}_i(\tilde{\mathbf{T}}_i^t,\mathbf{T}_i^t)
\right)\\
&+
\sum_{(p,c)\in\mathcal{E}_t}
\mathcal{L}\left(
\sqrt{\lambda_r}\,
\mathbf{r}_c\left(
\bar{\mathbf{T}}_{pc},
(\mathbf{T}_p^t)^{-1}\mathbf{T}_c^t
\right)
\right),
\end{aligned}
\label{eq:graph_pose_optimization}
\end{equation}
where $\mathcal{L}$ denotes the element-wise
robust soft-$L_1$ penalty.
The observation term maintains fidelity to
the initial pose estimates, while the relation
term penalizes deviations from persistent
rigid configurations.
We denote the optimized poses by
$\hat{\mathbf{T}}_i^t$, retaining
$\hat{\mathbf{T}}_i^t=\tilde{\mathbf{T}}_i^t$
for components not optimized.
Optimization is performed independently at
each observed keyframe without explicit
temporal smoothing.

\paragraph{Gauge-Consistent Motion Correction.}
Since 4DPM represents primitive motion in
component-specific gauge frames, the refined
poses must be transferred without altering
the original reference geometry.
Let $a_i$ denote the first observed keyframe
of component $i$.
We compute an anchor-relative correction
\begin{equation}
\mathbf{H}_i^t =
\left[
\hat{\mathbf{T}}_i^t
(\hat{\mathbf{T}}_i^{a_i})^{-1}
\right]
\left[
\tilde{\mathbf{T}}_i^t
(\tilde{\mathbf{T}}_i^{a_i})^{-1}
\right]^{-1}.
\label{eq:gauge_correction}
\end{equation}
This correction preserves the initial configuration
($\mathbf{H}_i^{a_i}=\mathbf{I}$) while updating
the component's relative motion.
We transfer the corrected motion to the original
4DPM transformations, preserving the geometry
and internal structure of the reconstructed primitives.

\subsection{Graph-Guided Motion Propagation}
\label{sec:motion_propagation}

\paragraph{Occlusion-Aware Relation Activation.}
The inferred persistent rigid relations enable
motion propagation from visible components to
their occluded rigid partners.
At each keyframe, we construct active rigid groups
from the verified physical-relation graph.
A relation remains active during occlusion
if its most recent joint observation supports
rigid coupling, while contradictory reobservations
prevent further propagation through that relation.
For each occluded component, we select a visible
driver within its active rigid group, prioritizing
relations with stronger geometric and temporal evidence.

\paragraph{Relative Motion Propagation.}
Given an occluded component $h$ and its visible
driver $d$, we transfer the driver's motion
relative to a jointly observed anchor keyframe $a$:
\begin{equation}
\mathbf{T}_h^{t,*} =
\mathbf{T}_d^{t,*}
(\mathbf{T}_d^{a,*})^{-1}
\mathbf{T}_h^{a,*},
\label{eq:graph_motion_propagation}
\end{equation}
where $\mathbf{T}^{t,*}$ denotes a refined or
propagated component pose.
This transfers the driver's anchor-to-current
motion to the occluded component, preserving
their reference relative configuration.
Upon reappearance, observed components use
their refined poses rather than propagated estimates.
If no valid driver is available, the component is not propagated.
\section{Experiments}

\subsection{Evaluation Protocol}
\label{sec:evaluation_protocol}

\paragraph{OmniGibson-Occlusion Benchmark.}
We construct OmniGibson-Occlusion using the
OmniGibson simulator~\cite{li2023behavior1k}, a synthetic benchmark
of six sequences covering diverse occlusion conditions (Fig.~\ref{fig:supp_dataset}).
The refrigerator and drawer sequences involve rotational
and translational motion, respectively, while Tray
examines temporary occlusion followed by reappearance.
Storage Box evaluates complete occlusion of objects
moving rigidly with their container.
Metric ground-truth geometry and motion remain available
throughout each sequence.

\paragraph{Multi-Object Benchmark.}
We also evaluate on the real-world Multi-Object benchmark
introduced by 4DPM~\cite{mazur2026primitive}, comprising seven sequences involving
object manipulation and multi-object interactions.
The dataset was captured using four synchronized cameras,
with a single monocular RGB stream used as input and
multi-view reconstructions providing pseudo-ground-truth
geometry.

\paragraph{Reconstruction Evaluation.}
We compare against $\pi^3$~\cite{wang2026pi3}, St4RTrack~\cite{feng2025st4rtrack}, POMATO~\cite{zhang2025pomato},
TraceAny~\cite{liu2026traceanything}, D4RT~	\cite{zhang2026d4rt}, and 4DPM~\cite{mazur2026primitive} on both benchmarks.
Following 4DPM, $\pi^3$ (Last-view Oracle)
uses only the final-frame pointmap, while
$\pi^3$ aggregates pointmaps without
motion compensation.
We warp reconstructed observations to the final
keyframe and aggregate them into a point cloud.
A separate Umeyama~\cite{umeyama1991least}
Sim(3) alignment is estimated between each
method's final-keyframe pointmap and the
corresponding ground-truth geometry.
We report precision, recall, and F-score
at a 1\,cm distance threshold on dynamic
scene geometry, averaged across sequences.

\paragraph{Relative Pose Evaluation.}
We evaluate object-level relative pose errors
on OmniGibson-Occlusion using metric ground truth.
Reconstructed components are associated with
ground-truth instances and aggregated to estimate object-level poses
through Kabsch alignment~\cite{kabsch1976solution}.
We evaluate moving objects relative to their
supporting or enclosing structures, excluding
the structures themselves.
Final-keyframe translation and rotation errors
are reported in centimeters and degrees.

\paragraph{Rigid-Relation Evaluation.}
We evaluate inferred fixed relations on
OmniGibson-Occlusion against ground-truth
rigid-coupling annotations.
Articulated connections and temporary co-motion
are not considered fixed rigid relations.
Ground-truth annotations are used solely for
evaluation.
We compute relation precision, recall, and F1
from pooled TP, FP, and FN across six sequences,
containing 17 ground-truth positive relations.

\paragraph{Implementation.}
We adopt the 4DPM visual frontend and extend its
reconstruction backend with physical-relation inference,
pose refinement, and motion propagation. All module ablations in Tab.~\ref{tab:component_ablation}
share the same initial reconstruction, while graph-based
variants use identical inferred relations.
For Tab.~\ref{tab:alternative_approaches},
we evaluate temporal smoothing configurations
across three window sizes and four blending strengths,
reporting the one with the highest reconstruction
F-score.
The 4DPM grouping baseline converts its original
motion-extension associations into rigid relations
and applies the same refinement and propagation modules.
The graph ablation in Tab.~\ref{tab:graph_ablation}
varies persistence classification while keeping
refinement and propagation unchanged.
All experiments use a single NVIDIA L40S GPU.
Detailed algorithmic procedures, implementation
parameters, and runtime analysis are provided
in the supplementary material.

\begin{table}[t]
    \centering
    \caption{\textbf{Ablation study of pose refinement and
    motion propagation.} We independently evaluate the two modules using
    the same initial reconstruction and inferred relations.
    Original denotes the 4DPM motion extension. 
    Best results are in bold and second-best results are underlined.}
    \label{tab:component_ablation}
    \resizebox{\columnwidth}{!}{
        \begin{tabular}{l|cc|ccc}
            \toprule
            \textbf{Method} &
            \textbf{Pose Refinement} &
            \textbf{Propagation} &
            \textbf{Mean F-score $\uparrow$} &
            \textbf{Precision $\uparrow$} &
            \textbf{Recall $\uparrow$} \\
            \midrule
            4DPM
                & -- & Original
                & 0.5864 & 0.5922 & 0.5843 \\
            + Pose Refinement
                & \checkmark & Original
                & 0.5981 & 0.5978 & 0.6019 \\
            + Graph Propagation
                & -- & Graph-Guided
                & \underline{0.6071} & \textbf{0.6140} & \underline{0.6042} \\
            \textbf{PRC4D (Ours)}
                & \checkmark & Graph-Guided
                & \textbf{0.6092} & \underline{0.6083} & \textbf{0.6140} \\
            \bottomrule
        \end{tabular}
    }
\end{table}

\begin{table}[t]
    \centering
    \caption{
    \textbf{Comparison with alternative approaches.}
    We compare our physical relation-constrained
    reconstruction with temporal SE(3) smoothing
    and rigid relations derived from 4DPM grouping.
    }
    \label{tab:alternative_approaches}
    \resizebox{\columnwidth}{!}{
        \begin{tabular}{l|ccc}
            \toprule
            \textbf{Method} &
            \textbf{Mean F-score $\uparrow$} &
            \textbf{Rel. Trans. ($cm$) $\downarrow$} &
            \textbf{Rel. Rot. ($^\circ$) $\downarrow$} \\
            \midrule
            4DPM
                & 0.5864 & 11.81 & 43.46 \\
            + Temporal SE(3) Smoothing
                & 0.5553 & 8.14 & 35.41 \\
            + 4DPM Grouping
                & 0.5668 & 13.70 & 18.93 \\
            \textbf{PRC4D (Ours)}
                & \textbf{0.6092}
                & \textbf{2.62}
                & \textbf{11.07} \\
            \bottomrule
        \end{tabular}
    }
\end{table}

\subsection{Persistent Reconstruction under Occlusion}
\label{sec:occlusion_results}

\paragraph{Geometry Persistence and Spatial Consistency.}
Tab.~\ref{tab:synthetic} reveals the distinction
between preserving observed geometry and maintaining
its correct spatial configuration.
While the last-view $\pi^3$ lacks previously observed
geometry, aggregating unaligned observations introduces
spatial inconsistencies.
4DPM preserves accumulated geometry, but independently
estimated component motions can still distort its
configuration.
Our method improves F-score across all six scenes
without modifying the underlying primitives,
highlighting the importance of consistent object motion.

\paragraph{Reconstruction under Changing Visibility.}
Fig.~\ref{fig:synthetic_qualitative} illustrates
temporary occlusion and reappearance in Tray,
and complete occlusion during rigid motion
in Storage Box.
The latter requires hidden components to follow
the motion of their visible rigid partner.
Our inferred relations preserve coherent relative
configurations across these changing visibility
conditions.

\subsection{Multi-Object Reconstruction and Physical Relations}
\label{sec:multiobject_results}

\paragraph{Reconstruction under Complex Motion.}
Tab.~\ref{tab:multiobject} evaluates our method
under diverse real-world object motions.
Unlike the consistent improvements observed in
synthetic occlusion scenarios, the gains vary
across multi-object sequences.
The larger improvements in MultiObj2 and Spin2
contrast with the unchanged result in MultiObj1
and the degradation in BallPush.
This suggests that physical-relation constraints
are beneficial when persistent rigid
couplings can be identified and visual
observations become incomplete or unreliable,
rather than acting as a general regularizer
for arbitrary object motion.
Additional quantitative and qualitative results
on HO3D~\cite{hampali2020honnotate} are provided in the supplementary material.

\paragraph{Persistent and Transient Interactions.}
Fig.~\ref{fig:multiobject_relations} illustrates
the distinction between persistent rigid couplings
and temporary interactions in a real-world sequence.
Objects may exhibit contact or similar motion over
a short interval without remaining rigidly attached.
Enforcing such relations throughout the sequence
would incorrectly constrain their subsequent
independent motions.
Our method therefore applies pose refinement and
motion propagation only to relations verified as
persistent, while excluding transient interactions
from the reconstruction constraints.

\begin{table}[t]
    \centering
    \caption{
    \textbf{Ablation of physical-relation graph construction.}
    We evaluate reconstruction quality and
    rigid-relation inference as temporal verification
    and graph-level consistency checks are
    progressively introduced.
    }
    \label{tab:graph_ablation}
    \resizebox{\columnwidth}{!}{
        \begin{tabular}{l|cc|cc}
            \toprule
            & \multicolumn{2}{c|}{\textbf{Reconstruction}} &
              \multicolumn{2}{c}{\textbf{Relation Inference}} \\
            \cmidrule(lr){2-3}
            \cmidrule(lr){4-5}
            \textbf{Method} &
            \textbf{Mean F-score $\uparrow$} &
            \textbf{Precision $\uparrow$} &
            \textbf{F1 $\uparrow$} &
            \textbf{Precision $\uparrow$} \\
            \midrule
            Contact + Motion Only
                & 0.4778 & 0.4671 & 0.8947 & 0.8095 \\
            + Temporal Verification
                & 0.6036 & 0.6043 & 0.9444 & 0.8947 \\
            \textbf{PRC4D (Ours)}
                & \textbf{0.6092}
                & \textbf{0.6083}
                & \textbf{0.9714}
                & \textbf{0.9444} \\
            \bottomrule
        \end{tabular}
    }
\end{table}

\subsection{Analysis of Physical Relation Constraints}
\label{sec:ablation}

\paragraph{Roles of Pose Refinement and Propagation.}
Tab.~\ref{tab:component_ablation} examines the
individual contributions of pose refinement
and graph-guided motion propagation.
Replacing 4DPM's original motion extension with
graph-guided propagation yields a larger improvement
than applying pose refinement alone.
This suggests that persistent physical relations
provide more reliable motion transfer through
occlusion than the original motion grouping.
Pose refinement further improves reconstruction
by correcting inconsistent observed poses.
Combining both modules achieves the highest
F-score and recall, maintaining coherent
object configurations across changing visibility.

\paragraph{Temporal Smoothness versus Relational Constraints.}
Tab.~\ref{tab:alternative_approaches} investigates
whether temporal smoothing or 4DPM
grouping can substitute for persistent physical
relations.
Independent temporal smoothing reduces relative
pose errors but degrades reconstruction F-score,
indicating that smoother individual trajectories
do not necessarily yield more accurate scene
reconstructions.
The complete evaluation across all smoothing
configurations is reported in the
supplementary material.

We further replace our inferred relations with
the original 4DPM motion-extension groups while
retaining the same refinement and propagation modules.
Despite improving rotational accuracy, this
alternative increases translation error and reduces
reconstruction F-score.
This suggests that enforcing rigid constraints
alone is insufficient when the underlying
component associations are unreliable.
Together, these results highlight the importance
of identifying reliable physical relations,
rather than merely smoothing or coupling
component motions.

\paragraph{Importance of Persistent Relation Verification.}
Tab.~\ref{tab:graph_ablation} shows that locally
plausible rigid relations can degrade reconstruction
when enforced without sufficient verification.
Temporal verification and graph-level consistency checks
progressively reduce false-positive relations from four to
one, while correctly identifying all 17 evaluated
ground-truth rigid relations.
These improvements increase relation precision
from 80.95\% to 94.44\%, accompanied by
substantial reconstruction gains.
The results demonstrate that the reliability
of inferred relations is critical when using
them as persistent motion constraints.

\section{Limitations}

Our method inherits the piecewise-rigid assumption
of 4DPM and cannot model complex non-rigid
deformations.
Relation inference depends on initial geometry
and motion estimates, potentially coupling
unattached components with similar motion.
Although articulated components can be reconstructed
independently, our constraints only handle
persistent fixed couplings.
Finally, motion propagation requires a visible
rigid partner, and sequence-level verification
requires the full video, limiting incremental
reconstruction.

\section{Conclusion}

We presented PRC4D, a framework that leverages
persistent rigid relations to maintain spatial
consistency across changing visibility.
By verifying physical relations through geometric
and temporal evidence, our method refines observed
component poses and propagates motion through
occlusion without altering the underlying geometry.
Experiments on synthetic occlusion and real-world
multi-object benchmarks demonstrate improved
reconstruction accuracy.
Our ablations further highlight that reliable
relation verification is essential to prevent
incorrect motion constraints from degrading
persistent reconstruction.

\section*{LLM Usage Statement}

OpenAI's ChatGPT was used to assist with manuscript,
language editing, and improving the clarity
and organization of technical explanations.
It was also used to discuss research framing,
experimental design, and the interpretation of
ablation results.
Anthropic's Claude was used to assist with
the design, generation, and refinement of
research figure.
The authors reviewed the generated suggestions
and retain full responsibility for all technical
claims, experimental results, and scientific conclusions.

{
    \small
    \bibliographystyle{ieeenat_fullname}
    \bibliography{main}

@String(CVPR= {IEEE Conf. Comput. Vis. Pattern Recog.})

@String(ICCV= {Int. Conf. Comput. Vis.})

@String(ECCV= {Eur. Conf. Comput. Vis.})

@String(ICLR = {Int. Conf. Learn. Represent.})

@String(CVPR  = {CVPR})

@String(ICCV  = {ICCV})

@String(ECCV  = {ECCV})

@String(ICLR  = {ICLR})

@inproceedings{mazur2026primitive,
  title = {{4D} Primitive-M{\^a}ch{\'e}: Glueing Primitives for Persistent {4D} Scene Reconstruction},
  author = {Mazur, Kirill and Taher, Marwan and Davison, Andrew J.},
  booktitle = {CVPR}, year = {2026}, pages = {7372--7381}
}

@inproceedings{wang2026pi3,
  title = {{$\pi^3$}: Permutation-Equivariant Visual Geometry Learning},
  author = {Wang, Yifan and Zhou, Jianjun and Zhu, Haoyi and Chang, Wenzheng and Zhou, Yang and Li, Zizun and Chen, Junyi and Pang, Jiangmiao and Shen, Chunhua and He, Tong},
  booktitle = {ICLR}, year = {2026}
}

@inproceedings{wang2025vggt,
  title = {{VGGT}: Visual Geometry Grounded Transformer},
  author = {Wang, Jianyuan and Chen, Minghao and Karaev, Nikita and Vedaldi, Andrea and Rupprecht, Christian and Novotny, David},
  booktitle = {CVPR}, year = {2025}
}

@inproceedings{wang2024dust3r,
  title = {{DUSt3R}: Geometric {3D} Vision Made Easy},
  author = {Wang, Shuzhe and Leroy, Vincent and Cabon, Yohann and Chidlovskii, Boris and Revaud, Jerome},
  booktitle = {CVPR}, year = {2024}, pages = {20697--20709}
}

@inproceedings{leroy2024mast3r,
  title = {Grounding Image Matching in {3D} with {MASt3R}},
  author = {Leroy, Vincent and Cabon, Yohann and Revaud, Jerome},
  booktitle = {ECCV}, year = {2024}
}

@inproceedings{murai2025mast3rslam,
  title = {{MASt3R-SLAM}: Real-Time Dense {SLAM} with {3D} Reconstruction Priors},
  author = {Murai, Riku and Dexheimer, Eric and Davison, Andrew J.},
  booktitle = {CVPR}, year = {2025}
}

@inproceedings{zhang2025monst3r,
  title = {{MonST3R}: A Simple Approach for Estimating Geometry in the Presence of Motion},
  author = {Zhang, Junyi and Herrmann, Charles and Hur, Junhwa and Jampani, Varun and Darrell, Trevor and Cole, Forrester and Sun, Deqing and Yang, Ming-Hsuan},
  booktitle = {ICLR}, year = {2025}
}

@inproceedings{li2025megasam,
  title = {{MegaSaM}: Accurate, Fast and Robust Structure and Motion from Casual Dynamic Videos},
  author = {Li, Zhengqi and Tucker, Richard and Cole, Forrester and Wang, Qianqian and Jin, Linyi and Ye, Vickie and Kanazawa, Angjoo and Holynski, Aleksander and Snavely, Noah},
  booktitle = {CVPR}, year = {2025}, pages = {10486--10496}
}

@inproceedings{feng2025st4rtrack,
  title = {{St4RTrack}: Simultaneous {4D} Reconstruction and Tracking in the World},
  author = {Feng, Haiwen and Zhang, Junyi and Wang, Qianqian and Ye, Yufei and Yu, Pengcheng and Black, Michael J. and Darrell, Trevor and Kanazawa, Angjoo},
  booktitle = {ICCV}, year = {2025}, pages = {8503--8513}
}

@inproceedings{zhang2025pomato,
  title = {{POMATO}: Marrying Pointmap Matching with Temporal Motions for Dynamic {3D} Reconstruction},
  author = {Zhang, Songyan and Ge, Yongtao and Tian, Jinyuan and Xu, Guangkai and Chen, Hao and Lv, Chen and Shen, Chunhua},
  booktitle = {ICCV}, year = {2025}, pages = {5680--5689}
}

@inproceedings{zhang2026d4rt,
  title = {Efficiently Reconstructing Dynamic Scenes One {D4RT} at a Time},
  author = {Zhang, Chuhan and Le Moing, Guillaume and Koppula, Skanda and Rocco, Ignacio and Momeni, Liliane and Xie, Junyu and Sun, Shuyang and Sukthankar, Rahul and Barral, Joelle K. and Hadsell, Raia and Ghahramani, Zoubin and Zisserman, Andrew and Zhang, Junlin and Sajjadi, Mehdi S. M.},
  booktitle = {CVPR}, year = {2026}, pages = {7382--7392}
}

@inproceedings{liu2026traceanything,
  title = {Trace Anything: Representing Any Video in {4D} via Trajectory Fields},
  author = {Liu, Xinhang and Xiao, Yuxi and Chen, Donny Y. and Feng, Jiashi and Tai, Yu-Wing and Tang, Chi-Keung and Kang, Bingyi},
  booktitle = {ICLR}, year = {2026}
}

@inproceedings{sucar2025dynamicpointmaps,
  title = {Dynamic Point Maps: A Versatile Representation for Dynamic {3D} Reconstruction},
  author = {Sucar, Edgar and Lai, Zihang and Insafutdinov, Eldar and Vedaldi, Andrea},
  booktitle = {ICCV}, year = {2025}
}

@inproceedings{wang2025shapeofmotion,
  title = {Shape of Motion: {4D} Reconstruction from a Single Video},
  author = {Wang, Qianqian and Ye, Vickie and Gao, Hang and Zeng, Weijia and Austin, Jake and Li, Zhengqi and Kanazawa, Angjoo},
  booktitle = {ICCV}, year = {2025}, pages = {9660--9672}
}

@inproceedings{wang2025c4d,
  title = {{C4D}: {4D} Made from {3D} through Dual Correspondences},
  author = {Wang, Shizun and Jiang, Zhenxiang and Yang, Xingyi and Wang, Xinchao},
  booktitle = {ICCV}, year = {2025}, pages = {7570--7580}
}

@inproceedings{wang2026complet4r,
  title = {{Complet4R}: Geometric Complete {4D} Reconstruction},
  author = {Wang, Weibang and Li, Kenan and Chen, Zhuoguang and Yuan, Yijun and Zhao, Hang},
  booktitle = {CVPR}, year = {2026}, pages = {341--351}
}

@inproceedings{gokmen2026com4d,
  title = {Inferring Compositional {4D} Scenes without Ever Seeing One},
  author = {Gokmen, Ahmet Berke and Chhatkuli, Ajad and Van Gool, Luc and Paudel, Danda Pani},
  booktitle = {CVPR}, year = {2026}
}

@inproceedings{lei2025mosca,
  title = {{MoSca}: Dynamic Gaussian Fusion from Casual Videos via {4D} Motion Scaffolds},
  author = {Lei, Jiahui and Weng, Yijia and Harley, Adam W. and Guibas, Leonidas and Daniilidis, Kostas},
  booktitle = {CVPR}, year = {2025}, pages = {6165--6177}
}

@inproceedings{xie2026cari4d,
  title = {{CARI4D}: Category Agnostic {4D} Reconstruction of Human-Object Interaction},
  author = {Xie, Xianghui and Wen, Bowen and Chang, Yan and Rabeti, Hesam and Li, Jiefeng and Yuan, Ye and Pons-Moll, Gerard and Birchfield, Stan},
  booktitle = {CVPR}, year = {2026}, pages = {14006--14016}
}

@inproceedings{li2026hat4d,
  title = {{HAT-4D}: Lifting Monocular Video for {4D} Multi-Object Interactions via Human-Agent Collaboration},
  author = {Li, Jiaxin and Wu, Yuxiang and Zhang, Zhenkai and Shi, Xinrui and Wang, Haoyuan and Zhao, Yichen and Su, Linxiang and Yu, Chenyang and Zhang, Mingyu and Ding, Yifan and Wen, Boran and Zhang, Li and Liu, Ruiyang and Li, Yong-Lu},
  booktitle = {ECCV}, year = {2026}, eprint = {2606.28215}, archivePrefix = {arXiv}
}

@inproceedings{zhou2026page4d,
  title = {{PAGE-4D}: Disentangled Pose and Geometry Estimation for {VGGT-4D} Perception},
  author = {Zhou, Kaichen and Wang, Yuhan and Chen, Grace and Beaudouin, Gaspard and Zhan, Fangneng and Liang, Paul and Wang, Mengyu},
  booktitle = {ICLR}, year = {2026}
}

@inproceedings{ost2021neural_scene_graphs,
  title = {Neural Scene Graphs for Dynamic Scenes},
  author = {Ost, Julian and Mannan, Fahim and Thuerey, Nils and Knodt, Julian and Heide, Felix},
  booktitle = {CVPR}, year = {2021}, pages = {2856--2865}
}

@inproceedings{zhang2022structuremotion,
  title = {Structure and Motion from Casual Videos},
  author = {Zhang, Zhoutong and Cole, Forrester and Li, Zhengqi and Snavely, Noah and Rubinstein, Michael and Freeman, William T.},
  booktitle = {ECCV}, year = {2022}, pages = {20--37}
}

@article{kumar2021superpixelsoup,
  title = {Superpixel Soup: Monocular Dense {3D} Reconstruction of a Complex Dynamic Scene},
  author = {Kumar, Suryansh and Dai, Yuchao and Li, Hongdong},
  journal = {IEEE Transactions on Pattern Analysis and Machine Intelligence},
  volume = {43}, number = {5}, pages = {1705--1717}, year = {2021},
  doi = {10.1109/TPAMI.2019.2955131}
}

@inproceedings{costeira1995multibody,
  title = {A Multi-Body Factorization Method for Motion Analysis},
  author = {Costeira, Joao and Kanade, Takeo},
  booktitle = {ICCV}, year = {1995}
}

@inproceedings{tresadern2005articulated,
  title = {Articulated Structure from Motion by Factorization},
  author = {Tresadern, Phil and Reid, Ian},
  booktitle = {CVPR}, year = {2005}
}

@inproceedings{luiten2024dynamic3dgaussians,
  title = {Dynamic {3D} Gaussians: Tracking by Persistent Dynamic View Synthesis},
  author = {Luiten, Jonathon and Kopanas, Georgios and Leibe, Bastian and Ramanan, Deva},
  booktitle = {International Conference on 3D Vision}, year = {2024}, doi = {10.1109/3DV62453.2024.00044}
}

@inproceedings{liu2023paris,
  title = {{PARIS}: Part-Level Reconstruction and Motion Analysis for Articulated Objects},
  author = {Liu, Jiayi and Mahdavi-Amiri, Ali and Savva, Manolis},
  booktitle = {ICCV}, year = {2023}, pages = {352--363}
}

@inproceedings{jiang2022ditto,
  title = {Ditto: Building Digital Twins of Articulated Objects from Interaction},
  author = {Jiang, Zhenyu and Hsu, Cheng-Chun and Zhu, Yuke},
  booktitle = {CVPR}, year = {2022}, pages = {5616--5626}
}

@inproceedings{harley2025alltracker,
  title = {{AllTracker}: Efficient Dense Point Tracking at High Resolution},
  author = {Harley, Adam W. and You, Yang and Sun, Xinglong and Zheng, Yang and Raghuraman, Nikhil and Gu, Yunqi and Liang, Sheldon and Chu, Wen-Hsuan and Dave, Achal and You, Suya and Ambrus, Rares and Fragkiadaki, Katerina and Guibas, Leonidas},
  booktitle = {ICCV}, year = {2025}, pages = {5253--5262}
}

@inproceedings{wang2024searaft,
  title = {{SEA-RAFT}: Simple, Efficient, Accurate {RAFT} for Optical Flow},
  author = {Wang, Yihan and Lipson, Lahav and Deng, Jia},
  booktitle = {ECCV}, year = {2024}
}

@inproceedings{karaev2024cotracker,
  title = {{CoTracker}: It Is Better to Track Together},
  author = {Karaev, Nikita and Rocco, Ignacio and Graham, Benjamin and Neverova, Natalia and Vedaldi, Andrea and Rupprecht, Christian},
  booktitle = {ECCV}, year = {2024}
}

@inproceedings{doersch2023tapir,
  title = {{TAPIR}: Tracking Any Point with Per-Frame Initialization and Temporal Refinement},
  author = {Doersch, Carl and Yang, Yi and Vecerik, Mel and Gokay, Dilara and Gupta, Ankush and Aytar, Yusuf and Carreira, Joao and Zisserman, Andrew},
  booktitle = {ICCV}, year = {2023}, pages = {10061--10072}
}

@inproceedings{li2023behavior1k,
  title = {{BEHAVIOR-1K}: A Benchmark for Embodied {AI} with 1,000 Everyday Activities and Realistic Simulation},
  author = {Li, Chengshu and Zhang, Ruohan and Wong, Josiah and Gokmen, Cem and Srivastava, Sanjana and Martin-Martin, Roberto and Wang, Chen and Levine, Gabrael and Lingelbach, Michael and Sun, Jiankai and Anvari, Mona and Hwang, Minjune and Sharma, Manasi and Aydin, Arman and Bansal, Dhruva and Hunter, Samuel and Kim, Kyu-Young and Lou, Alan and Matthews, Caleb R. and Villa-Renteria, Ivan and Tang, Jerry Huayang and Tang, Claire and Xia, Fei and Savarese, Silvio and Gweon, Hyowon and Liu, Karen and Wu, Jiajun and Fei-Fei, Li},
  booktitle = {Proceedings of the Conference on Robot Learning},
  series = {Proceedings of Machine Learning Research}, volume = {205}, pages = {80--93}, year = {2023}
}

@article{umeyama1991least,
  title = {Least-Squares Estimation of Transformation Parameters Between Two Point Patterns},
  author = {Umeyama, Shinji},
  journal = {IEEE Transactions on Pattern Analysis and Machine Intelligence}, volume = {13}, number = {4}, pages = {376--380}, year = {1991}, doi = {10.1109/34.88573}
}

@article{kabsch1976solution,
  title = {A Solution for the Best Rotation to Relate Two Sets of Vectors},
  author = {Kabsch, Wolfgang},
  journal = {Acta Crystallographica Section A}, volume = {32}, number = {5}, pages = {922--923}, year = {1976}, doi = {10.1107/S0567739476001873}
}

@inproceedings{wang2025cut3r,
  title = {Continuous 3D Perception Model with Persistent State},
  author = {Wang, Qianqian and Zhang, Yifei and Holynski, Aleksander and Efros, Alexei A. and Kanazawa, Angjoo},
  booktitle = CVPR,
  pages = {10510--10522},
  year = {2025}
}

@inproceedings{sucar2026vdpm,
  title = {{V-DPM}: 4D Video Reconstruction with Dynamic Point Maps},
  author = {Sucar, Edgar and Insafutdinov, Eldar and Lai, Zihang and Vedaldi, Andrea},
  booktitle = CVPR,
  pages = {14502--14511},
  year = {2026}
}

@inproceedings{chen2025easi3r,
  title = {{Easi3R}: Estimating Disentangled Motion from {DUSt3R} Without Training},
  author = {Chen, Xingyu and Chen, Yue and Xiu, Yuliang and Geiger, Andreas and Chen, Anpei},
  booktitle = ICCV,
  pages = {9158--9168},
  year = {2025}
}

@inproceedings{wang2023omnimotion,
  title = {Tracking Everything Everywhere All at Once},
  author = {Wang, Qianqian and Chang, Yen-Yu and Cai, Ruojin and Li, Zhengqi and Hariharan, Bharath and Holynski, Aleksander and Snavely, Noah},
  booktitle = ICCV,
  pages = {19795--19806},
  year = {2023}
}

@inproceedings{matsuki2025_4dtam,
  title = {{4DTAM}: Non-Rigid Tracking and Mapping via Dynamic Surface Gaussians},
  author = {Matsuki, Hidenobu and Bae, Gwangbin and Davison, Andrew J.},
  booktitle = CVPR,
  pages = {26921--26932},
  year = {2025}
}

@inproceedings{huang2025vivid4d,
  title = {{Vivid4D}: Improving 4D Reconstruction from Monocular Video by Video Inpainting},
  author = {Huang, Jiaxin and Miao, Sheng and Yang, Bangbang and Ma, Yuewen and Liao, Yiyi},
  booktitle = ICCV,
  pages = {12592--12604},
  year = {2025}
}

@inproceedings{hampali2020honnotate,
  title={HOnnotate: A Method for 3D Annotation
         of Hand and Object Poses},
  author={Hampali, Shreyas and Rad, Mahdi
          and Oberweger, Markus and Lepetit, Vincent},
  booktitle={Proceedings of the IEEE/CVF Conference
             on Computer Vision and Pattern Recognition},
  year={2020}
}
}

\clearpage
\setcounter{page}{1}
\maketitlesupplementary

\setcounter{section}{0}
\setcounter{table}{0}
\setcounter{figure}{0}
\setcounter{algorithm}{0}

\renewcommand{\thetable}{S\arabic{table}}
\renewcommand{\thefigure}{S\arabic{figure}}
\renewcommand{\thealgorithm}{A\arabic{algorithm}}

\providecommand{\theHsection}{}
\providecommand{\theHtable}{}
\providecommand{\theHfigure}{}
\providecommand{\theHalgorithm}{}

\renewcommand{\theHsection}{supp.\arabic{section}}
\renewcommand{\theHtable}{supp.\arabic{table}}
\renewcommand{\theHfigure}{supp.\arabic{figure}}
\renewcommand{\theHalgorithm}{supp.\arabic{algorithm}}

\section{Implementation Details}

\paragraph{Algorithm overview.}
Alg.~\ref{alg:supp_prc4d} summarizes the detailed
PRC4D pipeline, from rigid-relation inference to
pose refinement and motion propagation.
Temporal verification distinguishes transient
contacts from ambiguous motion evidence, while
graph-level consistency resolves uncertain
relations using trusted rigid groups and
redundant support.

\paragraph{Parameter settings.}
Tab.~\ref{tab:supp_const} reports the key settings used for the
quantitative evaluations. The child of a candidate pair is chosen as the component
with the smaller median OBB volume, and its size $s$ is the median
OBB diagonal. Lengths are expressed in reconstruction units unless
scaled by $s$; the voxel size is expressed relative to the median
point spacing. The parameters $\alpha$, $b$, and $\beta$ correspond
to the motion-consistency criterion in Eq.~(2) of the main paper.
For observed keyframes, we estimate a robust reference relative
transformation for each verified relation and optimize component
poses using the observation and relation terms in Eq.~(4). For
occluded components, motion is propagated from a visible rigid
partner when one is available, while the original primitive
geometry remains unchanged.

\begin{algorithm}[t]
\caption{PRC4D: Relation-Constrained Persistent Reconstruction}
\label{alg:supp_prc4d}
\small
\begin{algorithmic}[1]

\Require 4DPM components $\mathcal{O}$,
initial poses $\tilde{\mathbf T}$,
and transforms $\mathbf U$
\Ensure Verified relations $\mathcal E$
and corrected poses $\mathbf T^*$

\Statex \textbf{Persistent Relation Inference}
\State $\mathcal P \gets \textsc{OBBCandidates}(\mathcal O)$
\State $\mathcal C \gets \emptyset$
\ForAll{$(i,j)\in\mathcal P$}
    \State $\mathcal I_{ij}\gets
    \textsc{ContactIntervals}(i,j)$
    \If{$\textsc{ValidContactMotion}(i,j,\mathcal I_{ij})$}
        \State $\mathcal C\gets\mathcal C\cup\{(i,j)\}$
        \State $b_{ij}\gets
        \textsc{TemporalVerification}(i,j)$
    \EndIf
\EndFor

\State $\mathcal G\gets
\textsc{BuildTrustedGroups}(\mathcal C,b)$
\State $\mathcal E\gets
\textsc{GraphConsistency}(\mathcal C,b,\mathcal G)$

\Statex \textbf{Graph-Constrained Pose Refinement}
\ForAll{$(p,c)\in\mathcal E$}
    \State $\bar{\mathbf T}_{pc}\gets
    \textsc{HuberMean}
    (\{(\tilde{\mathbf T}_p^t)^{-1}
    \tilde{\mathbf T}_c^t\}_{t\in\mathcal I_{pc}})$
\EndFor
\ForAll{observed keyframes $t$}
    \State $\hat{\mathbf T}^t\gets
    \textsc{RefinePoses}
    (\tilde{\mathbf T}^t,\mathcal E_t,\bar{\mathbf T})$
    \Comment{Eq.~(4)}
\EndFor
\State $\mathbf T^*\gets
\textsc{CorrectMotion}(\mathbf U,\tilde{\mathbf T},
\hat{\mathbf T})$
\Comment{Eq.~(5)}

\Statex \textbf{Graph-Guided Motion Propagation}
\ForAll{keyframes $t$ and occluded components $h$}
    \State $d\gets
    \textsc{VisibleDriver}(h,\mathcal E,t)$
    \If{$d\neq\emptyset$}
        \State $a\gets
        \textsc{SharedAnchor}(h,d,t)$
        \State $\mathbf T_h^{t,*}\gets
        \mathbf T_d^{t,*}
        (\mathbf T_d^{a,*})^{-1}
        \mathbf T_h^{a,*}$
        \Comment{Eq.~(6)}
    \EndIf
\EndFor

\State \Return $\mathcal E,\mathbf T^*$

\end{algorithmic}
\end{algorithm}

\begin{table}[!ht]\centering\small
\caption{\textbf{Key implementation parameters.} Fixed settings of physical-relation inference and pose refinement used in the quantitative evaluations. $s$ is the child size (median OBB diagonal); lengths are in reconstruction units unless scaled by $s$.}
\label{tab:supp_const}
\setlength{\tabcolsep}{3pt}
\begin{tabular}{lc}\toprule
\textbf{Parameter} & \textbf{Value} \\ \midrule
\multicolumn{2}{l}{\textbf{\emph{Contact evidence (Sec.~3.2)}}} \\
Voxel size ($\times$ median point spacing) & $5.0$ \\
Contact coverage: enter / exit & $0.01$ / $0.005$ \\
\midrule
\multicolumn{2}{l}{\textbf{\emph{Rigid motion consistency (Eq.~2)}}} \\
$\alpha$ ($\times s$) / $b$ / $\beta$ & $0.4$ / $0.08$ / $0.12$ \\
Consistency threshold on $\epsilon$ & $1.0$ \\
Measurable motion: transl.\ / rot.\ (rad) & $0.02$ / $0.05$ \\
\midrule
\multicolumn{2}{l}{\textbf{\emph{Persistence verification (Eq.~3)}}} \\
Terminal drift on $d$: threshold / last keyframes & $0.16$ / $2$ \\
Trusted-group seed: min.\ contact keyframes & $4$ \\
Redundant support: distinct neighbors & $4$ \\
\midrule
\multicolumn{2}{l}{\textbf{\emph{Pose refinement (Eq.~4)}}} \\
$\lambda_o$ / $\lambda_r$ & $1.0$ / $4.0$ \\
Huber scale: transl.\ ($\times s$) / rot. & $0.01$ / $1^\circ$ \\
\bottomrule\end{tabular}\end{table}
\begin{table*}[t]
\centering
\small
\setlength{\tabcolsep}{4pt}
\renewcommand{\arraystretch}{0.92}

\caption{
\textbf{Scene-wise module ablation on OmniGibson-Occlusion.}
Reconstruction F-score ($\uparrow$) for pose refinement,
graph-guided propagation, and their combination.
Best and second-best results are highlighted in bold
and underlined, respectively.
}
\label{tab:supp_module}

\resizebox{0.90\textwidth}{!}{%
\begin{tabular}{lcccccc|c}
\toprule
Method
& Fridge1 & Fridge2 & Drawer1 & Drawer2
& Tray & Storage Box & Mean \\
\midrule

4DPM
& \underline{0.5835} & 0.4872
& \underline{0.7687} & 0.6704
& 0.3772 & 0.6314 & 0.5864 \\

+ Pose Refinement
& \textbf{0.5982} & 0.4873
& \textbf{0.7950} & \underline{0.6766}
& 0.3853 & 0.6464 & 0.5981 \\

+ Graph Propagation
& \underline{0.5835} & \underline{0.5011}
& \underline{0.7687} & 0.6739
& \textbf{0.4098} & \textbf{0.7056}
& \underline{0.6071} \\

\midrule

\textbf{PRC4D (Ours)}
& \textbf{0.5982} & \textbf{0.5077}
& \textbf{0.7950} & \textbf{0.6802}
& \underline{0.3952} & \underline{0.6790}
& \textbf{0.6092} \\

\bottomrule
\end{tabular}%
}

\vspace{-3mm}
\end{table*}

\begin{table*}[t]
\centering
\small
\setlength{\tabcolsep}{4pt}
\renewcommand{\arraystretch}{0.92}

\caption{
\textbf{Scene-wise physical-relation graph ablation.}
Reconstruction F-score ($\uparrow$) and fixed-relation
TP/FP/FN on OmniGibson-Occlusion.
Best reconstruction scores and the lowest total
false-positive count are highlighted in bold.
}

\label{tab:supp_graph}

\resizebox{0.90\textwidth}{!}{%
\begin{tabular}{lcccccc|c}
\toprule
\textbf{Reconstruction F-score} $\uparrow$
& Fridge1 & Fridge2 & Drawer1 & Drawer2
& Tray & Storage Box & Mean \\
\midrule

Contact + Motion Only
& 0.3544 & 0.3209
& \textbf{0.7950} & \textbf{0.6802}
& 0.3951 & 0.3215 & 0.4778 \\

+ Temporal Verification
& \textbf{0.5982} & \textbf{0.5077}
& \textbf{0.7950} & \textbf{0.6802}
& \textbf{0.3952} & 0.6454 & 0.6036 \\

\textbf{PRC4D (Ours)}
& \textbf{0.5982} & \textbf{0.5077}
& \textbf{0.7950} & \textbf{0.6802}
& \textbf{0.3952} & \textbf{0.6790}
& \textbf{0.6092} \\

\midrule[\heavyrulewidth]

\textbf{Fixed relations (TP/FP/FN)}
& Fridge1 & Fridge2 & Drawer1 & Drawer2
& Tray & Storage Box & \textbf{Total FP $\downarrow$} \\
\midrule

Contact + Motion Only
& 4/1/0 & 2/2/0 & 3/0/0 & 3/0/0
& 3/0/0 & 2/1/0 & 4 \\

+ Temporal Verification
& 4/0/0 & 2/1/0 & 3/0/0 & 3/0/0
& 3/0/0 & 2/1/0 & 2 \\

\textbf{PRC4D (Ours)}
& 4/0/0 & 2/1/0 & 3/0/0 & 3/0/0
& 3/0/0 & 2/0/0 & \textbf{1} \\

\bottomrule
\end{tabular}%
}

\vspace{-3mm}
\end{table*}

\begin{table}[t]
\centering
\caption{
\textbf{Temporal SE(3) smoothing sweep on OmniGibson-Occlusion.}
We vary window size $W$ and blending coefficient
$\alpha_{\mathrm{sm}}$.
Best and second-best results among smoothing
configurations are highlighted in bold and
underlined, respectively.
The original 4DPM is shown for reference.
}
\label{tab:supp_smooth}

\resizebox{\columnwidth}{!}{%
\begin{tabular}{cc|ccc}
\toprule
\textbf{$W$} &
\textbf{$\alpha_{\mathrm{sm}}$} &
\textbf{F-score $\uparrow$} &
\textbf{Rel. Trans. (cm) $\downarrow$} &
\textbf{Rel. Rot. ($^\circ$) $\downarrow$} \\
\midrule

\multicolumn{2}{c|}{4DPM ($\alpha_{\mathrm{sm}}=0$)}
& 0.5864 & 11.81 & 43.46 \\

\midrule

3 & 0.25 & \textbf{0.5553} & 8.14 & 35.41 \\
3 & 0.50 & \underline{0.5421} & 9.19 & 24.48 \\
3 & 0.75 & 0.5349 & \underline{7.43} & \underline{23.42} \\
3 & 1.00 & 0.5220 & 8.62 & 26.59 \\

\midrule

5 & 0.25 & 0.5405 & 9.22 & 24.37 \\
5 & 0.50 & 0.5176 & 7.63 & 23.88 \\
5 & 0.75 & 0.4764 & 10.05 & 29.85 \\
5 & 1.00 & 0.4348 & 12.70 & 35.73 \\

\midrule

7 & 0.25 & 0.5268 & \textbf{6.68} & \textbf{23.35} \\
7 & 0.50 & 0.4786 & 8.92 & 30.95 \\
7 & 0.75 & 0.4161 & 11.48 & 36.72 \\
7 & 1.00 & 0.3713 & 15.40 & 38.16 \\

\bottomrule
\end{tabular}%
}
\end{table}

\begin{table}[t]
\centering
\caption{
\textbf{Runtime breakdown on OmniGibson-Occlusion.}
Average processing time per sequence on an NVIDIA L40S GPU.
Checkmarks indicate stages used by each method.
All times are rounded to the nearest second.
}
\label{tab:supp_runtime}

\setlength{\tabcolsep}{5pt}
\renewcommand{\arraystretch}{0.95}

\resizebox{\columnwidth}{!}{%
\begin{tabular}{l|cc|c}
\toprule
\textbf{Part}
& \textbf{4DPM}
& \textbf{PRC4D}
& \textbf{Time (s)} \\
\midrule

Segmentation (SAM2)
& $\checkmark$ & $\checkmark$ & 56 \\

$\pi^3$
& $\checkmark$ & $\checkmark$ & 2 \\

Correspondence (AllTracker)
& $\checkmark$ & $\checkmark$ & 3 \\

Frontend combined
& $\checkmark$ & $\checkmark$ & 60 \\

Backend
& $\checkmark$ & $\checkmark$ & 10 \\

\midrule

Motion segmentation (4DPM)
& $\checkmark$ & -- & 2 \\

Relation graph
& -- & $\checkmark$ & 93 \\

Pose refinement
& -- & $\checkmark$ & 3 \\

Graph-guided propagation
& -- & $\checkmark$ & 1 \\

\midrule

\textbf{4DPM total}
& $\checkmark$ & -- & \textbf{73} \\

\textbf{PRC4D total}
& -- & $\checkmark$ & \textbf{167} \\

\bottomrule
\end{tabular}%
}
\end{table}

\section{Additional Results on HO3D}
\paragraph{Evaluation protocol.}
HO3D contains real-world
hand-object interaction sequences involving object
manipulation, partial occlusion, and changing visibility.
We evaluate 85 clips of 150 frames from seven sequences
using camera~0, following the reconstruction protocol
in the main paper.

\paragraph{Quantitative results.}
Tab.~\ref{tab:supp_ho3d} evaluates PRC4D on
real-world hand-object interactions beyond
the main benchmarks.
PRC4D improves reconstruction F-score over 4DPM
across all seven sequences, while preserving
the original geometry.
These results suggest that relation-constrained
motion correction can improve spatial consistency
in different real-world settings.

\begin{figure*}[t]
    \centering
    \includegraphics[width=0.90\textwidth]
    {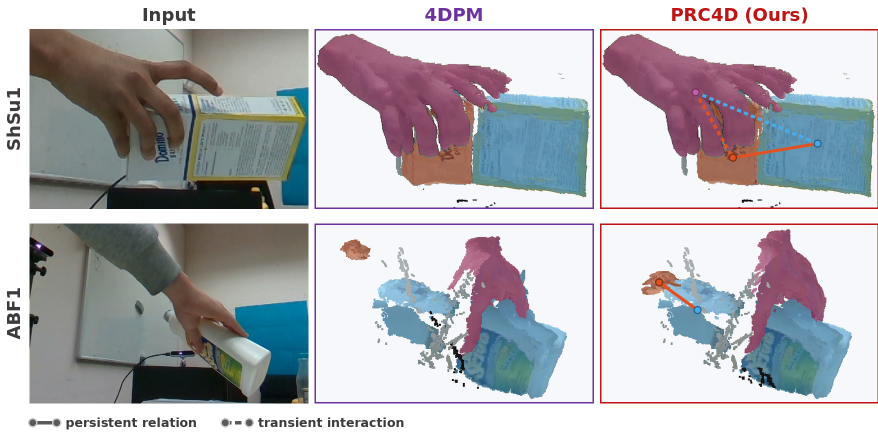}
    \caption{\textbf{Persistent physical relations on HO3D.}
    We compare 4DPM and ours on ShSu1 (top), where a box is reconstructed as two
    separate faces, and ABF1 (bottom), where the bottle cap is no longer observed.
    PRC4D links the box faces through a persistent relation while keeping hand
    contacts transient, and moves the unobserved cap with the bottle.
    Solid and dashed edges indicate persistent and transient relations, respectively.}
    \label{fig:supp_ho3d}
\end{figure*}

\paragraph{Qualitative comparisons.}
Fig.~\ref{fig:supp_ho3d} illustrates how PRC4D
maintains consistent configurations of
separately reconstructed components during
hand-object interactions.
By preserving persistent rigid relations while
excluding transient contacts, PRC4D reduces
component-level pose inconsistencies observed
in 4DPM, even when the components remain visible.

\begin{table*}[t]
\centering
\caption{
\textbf{Quantitative comparison on HO3D.}
We report F-score ($\uparrow$) for each of the seven sequences,
along with mean F-score, precision, and recall at a 1\,cm threshold.
Best and second-best results among non-oracle
methods are highlighted in bold and underlined,
respectively.
The last-view oracle uses only the final-frame
pointmap and is excluded from the ranking.
}
\label{tab:supp_ho3d}

\setlength{\tabcolsep}{4pt}
\renewcommand{\arraystretch}{0.92}

\resizebox{0.93\textwidth}{!}{%
\begin{tabular}{lccccccc|ccc}
\toprule
Method
& ABF1 & BB1 & GPMF1 & GSF1
& MDF1 & ShSu1 & SiBF1
& Mean & Precision & Recall \\
\midrule

$\pi^3$ (Last-view Oracle)
& 0.3806 & 0.4192 & 0.4180 & 0.4237
& 0.3706 & 0.4245 & 0.3913
& 0.4040 & 0.8405 & 0.2684 \\

\midrule

$\pi^3$
& 0.4583 & 0.5034 & 0.6632 & 0.5716
& 0.6017 & 0.7166 & 0.6663
& 0.5973 & 0.4956 & 0.7902 \\

St4RTrack
& 0.5953 & 0.5345 & 0.6146 & 0.6092
& 0.5978 & 0.6582 & 0.6413
& 0.6073 & 0.5158 & 0.7856 \\

POMATO
& 0.4770 & 0.4717 & 0.6497 & 0.4272
& 0.5787 & 0.7463 & 0.7117
& 0.5803 & 0.4922 & 0.7874 \\

TraceAny
& 0.5837 & 0.5756 & 0.7892 & 0.6696
& 0.6874 & 0.7031 & 0.6801
& 0.6698 & 0.5665 & \textbf{0.8545} \\

D4RT
& 0.5284 & 0.6985 & 0.7920 & 0.5330
& 0.6632 & 0.7990 & 0.7252
& 0.6770 & 0.6549 & 0.7372 \\

4DPM
& \underline{0.7965} & \underline{0.8161}
& \underline{0.7984} & \underline{0.7812}
& \underline{0.8010} & \underline{0.8125}
& \underline{0.7638}
& \underline{0.7957} & \underline{0.7584}
& \underline{0.8455} \\

\midrule

\textbf{PRC4D (Ours)}
& \textbf{0.8025} & \textbf{0.8220}
& \textbf{0.8162} & \textbf{0.7864}
& \textbf{0.8027} & \textbf{0.8141}
& \textbf{0.7659}
& \textbf{0.8014} & \textbf{0.7712}
& 0.8416 \\

\bottomrule
\end{tabular}%
}

\vspace{-3mm}
\end{table*}

\section{Scene-wise Ablation Studies}

\paragraph{Pose refinement and propagation.}
Tab.~\ref{tab:supp_module} shows that the two modules
contribute differently across scenes.
Graph-guided propagation is particularly effective
in Tray and Storage Box, where occluded components
must follow visible rigid partners.
Pose refinement improves reconstruction in scenes
such as Fridge1 and Drawer1 by correcting observed
component configurations.
Combining both modules yields the best overall
performance, although their benefits vary by scene.

\paragraph{Relation verification and graph consistency.}
Tab.~\ref{tab:supp_graph} demonstrates the
importance of rejecting unreliable rigid
relations.
In Fridge1, Fridge2, and Storage Box,
locally plausible but incorrect relations
substantially degrade reconstruction.
Temporal verification removes contradictory
relations, while graph-level consistency
further rejects incompatible couplings.
These results show that reliable relation
selection is essential for maintaining
coherent component configurations.

Among 25 initially annotated relation candidates,
we exclude five without reconstructed endpoints
and three that do not meet the fixed-coupling
evaluation criteria.
The remaining 17 relations are used for evaluation,
with metrics computed from pooled TP, FP, and FN.

\section{Temporal Smoothing Baseline}
Tab.~\ref{tab:supp_smooth} evaluates temporal
smoothing across 12 parameter configurations.
Although smoothing can reduce relative pose errors,
all tested configurations degrade reconstruction
F-score compared with 4DPM.
This suggests that improving temporal smoothness
alone is insufficient for reliable reconstruction,
highlighting the importance of enforcing
persistent relations between components.

\section{Additional OmniGibson-Occlusion Visualizations}

\begin{figure*}[t]
    \centering
    \includegraphics[width=1.00\textwidth]
    {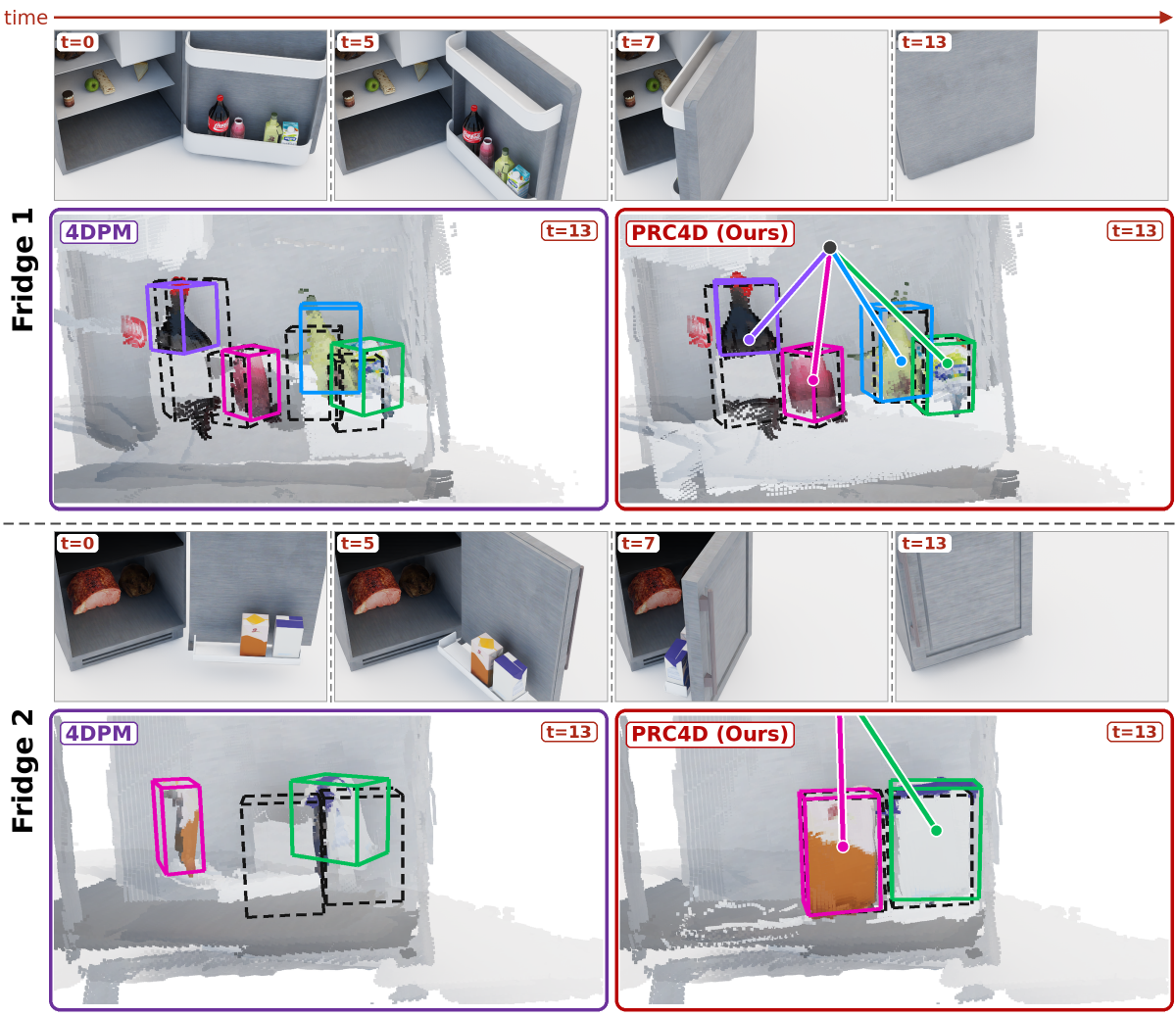}

    \caption{
    \textbf{Additional OmniGibson-Occlusion
    reconstructions: refrigerators.}
    We compare 4DPM and PRC4D on
    \textbf{Fridge1} (top) and
    \textbf{Fridge2} (bottom), where objects
    on the door shelves rotate with the
    closing door and become fully occluded.
    Persistent rigid relations reduce the
    drift of occluded objects observed in 4DPM.
    }
    \label{fig:supp_qual_fridge}
\end{figure*}

\begin{figure*}[t]
    \centering
    \includegraphics[width=1.00\textwidth]
    {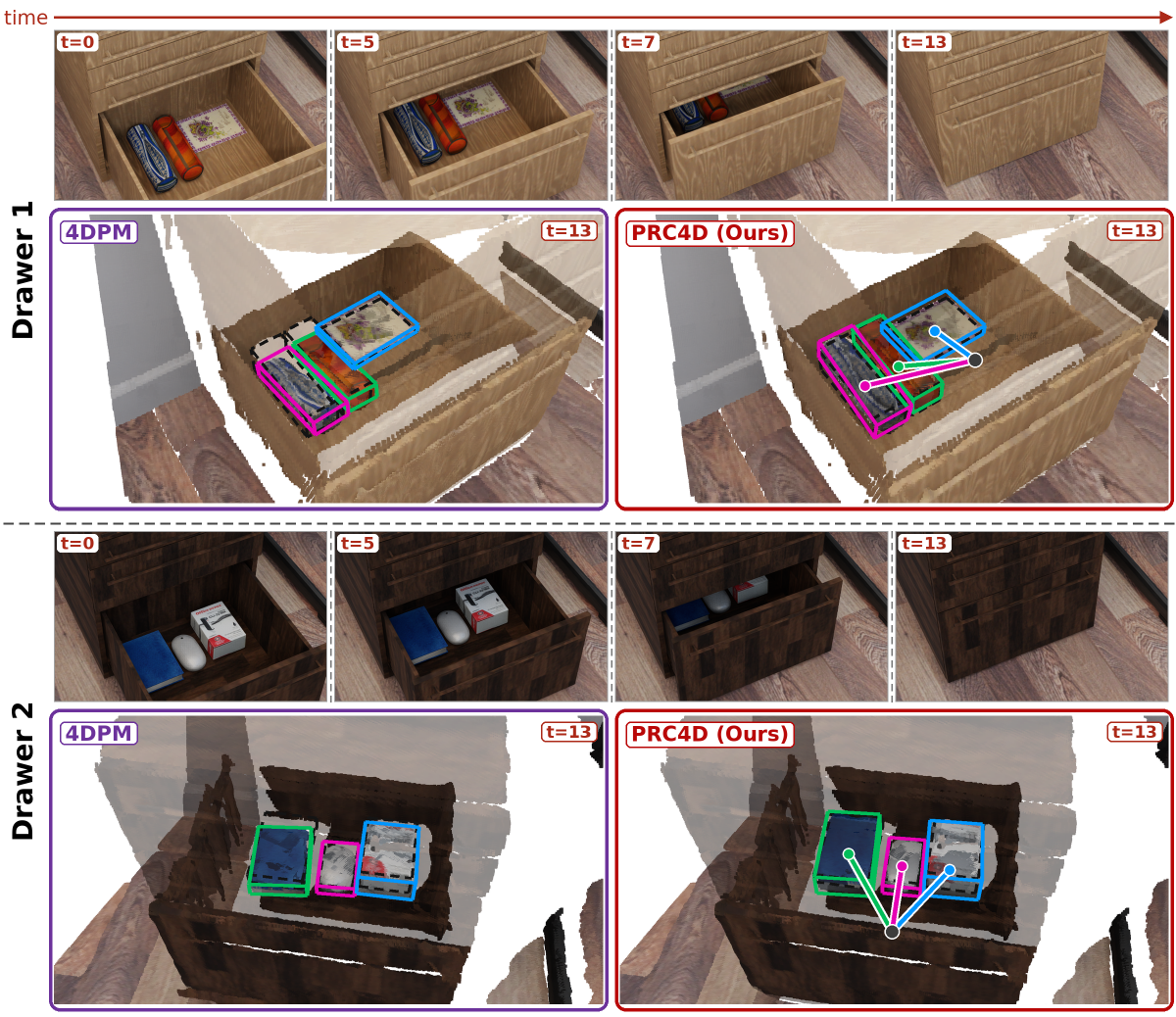}

    \caption{
    \textbf{Additional OmniGibson-Occlusion
    reconstructions: drawers.}
    We compare 4DPM and PRC4D on
    \textbf{Drawer1} (top) and
    \textbf{Drawer2} (bottom), where objects
    move with the closing drawer and
    become occluded.
    Persistent rigid relations allow hidden
    objects to follow the translating drawer.
    }
    \label{fig:supp_qual_drawer}
\end{figure*}

Figs.~\ref{fig:supp_qual_fridge} and
\ref{fig:supp_qual_drawer} further illustrate
the importance of maintaining relative object
configurations under occlusion.
While 4DPM preserves previously observed geometry,
its estimated component motions can lead to
inconsistent object placements.
PRC4D maintains spatial consistency through
verified rigid relations across both rotational
and translational motions.
The dashed reference boxes visualize relative
configuration drift rather than absolute pose errors.

\section{Runtime Analysis}

\paragraph{Runtime breakdown.}
Tab.~\ref{tab:supp_runtime} shows that
the additional computational cost of PRC4D
is dominated by physical-relation graph
construction, particularly geometric contact
evaluation between reconstructed components.
In contrast, pose refinement and graph-guided
motion propagation incur relatively little
overhead.
This identifies efficient relation inference
as the main opportunity for reducing runtime.

\begin{figure*}[p]\centering
\includegraphics[width=\textwidth]{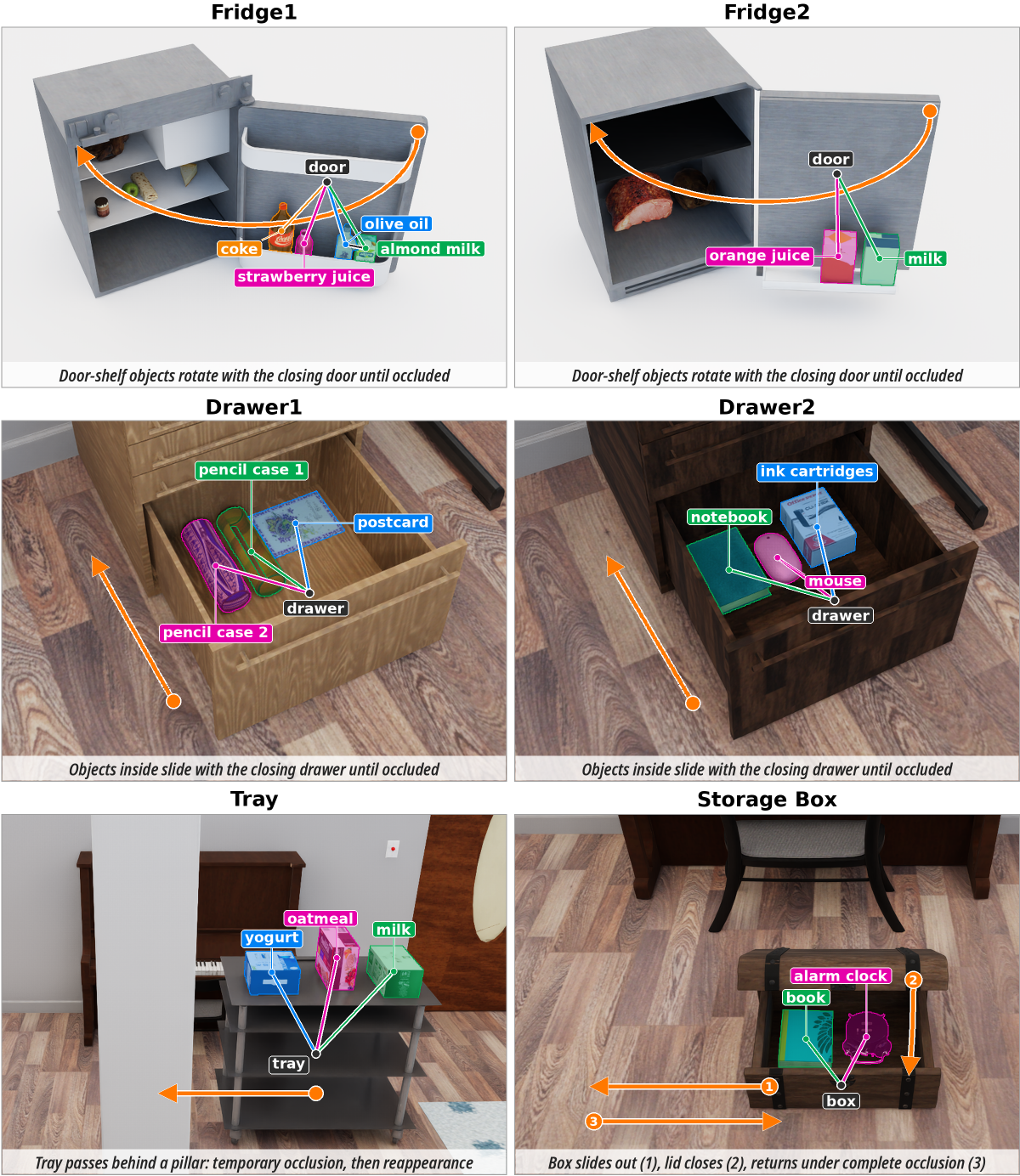}
\caption{
    \textbf{Overview of OmniGibson-Occlusion.}
    Six sequences cover rotational and translational
    motion, temporary occlusion, and complete occlusion.
    Orange arrows indicate object or container motion,
    while colored connections show ground-truth rigid
    relations between scene components.
    }
\label{fig:supp_dataset}
\end{figure*}

\end{document}